# Angry Birdsにおける建物構築文法による中華風と和風の建築物の自動生成
# Procedural Generation of Angry Birds Levels using Building Constructive Grammar with Chinese-Style and/or Japanese-Style Models

蒋 雨軒/国際語学学院，海段 美紗希/立命館大学 情報理工学部，朱 晉賢/立命館大学大学院 情報理工学研究科，原田 智宏，ターウォンマット ラック/立命館大学 情報理工学部

YuXuan Jiang/Interculture Language Academy, Misaki Kaidan/College of Information Science and Engineering (CISE), Ritsumeikan University, Chun Yin Chu/Graduate School of Information Science and Engineering, Ritsumeikan University, Tomohiro Harada, and Ruck Thawonmas*/CISE, Ritsumeikan University (*ruck@is.ritsumei.ac.jp)

Abstract: This paper presents a procedural generation method that creates visually attractive levels for the Angry Birds game. Besides being an immensely popular mobile game, Angry Birds has recently become a test bed for various artificial intelligence technologies. We propose a new approach for procedurally generating Angry Birds levels using Chinese style and Japanese style building structures. A conducted experiment confirms the effectiveness of our approach with statistical significance.

## 1. Introduction

This paper discusses how aesthetic Angry Birds level can be generated propocedurally. In our proposed method, Angry Birds level is created by inputting Chinese style or Japanese style strucutres into generation rules written in our proposed building constructive grammar. The generated Angry Birds level not only mimics Chinese and Japanese architectural styles, but also combines and mixes the features of both styles into a aesthetic structure. A survey was conducted to test if our approach could create levels using Chinese and Japanese architectural features that are recognizable to players. This research used the Angry Birds clone developed by Ferreira and Toledo (2014), which can be downloaded from GitHub[1].

## 2. Related Research

### 2.1. Generating buildings with model library

In existing research, generating Asian style 3D buildings using a model library and pre-defined parameters was conducted (Teoh, 2009). However, in order to generate a building with a desired style, the user has to set the parameters manually.

### 2.2. Video Game Description Language (VGDL)

VGDL (Ebner et al., 2013) is a language for describing game maps and game rules. VGDL is acclaimed for its readability, simplicity and scalability, and is being used to generate a wide range of games for the General Video Game AI Competition. However, the current version of VGDL does not support gravity or physics simulation, resulting in it being an unsuitable language for generating realistic buildings, which are subjected to physical constraints.

## 3. Proposed Method

To overcome the limits of VGDL, we propose a grammar called 2D Building Construction Grammar (2D-BCG). A level can be defined by a set of 2D-BCG rules, and rendered using structures with different styles. As a result, a structure with Chinese or Japanese architectural style can be generated automatically.

### 3.1. 2D Building Construction Grammar

2D-BCG is used for describing a building structure. In this set of rules, a building is divided into 3 parts: base, main and roofs, from bottom to top. Building is formed by 7 elements, including wall, floor, beam, window, door, roof and toproof (Fig. 1 left).

The 2D-BCG is written in Backus-Naur Form as shown below. In the following, variables enclosed in angle brackets (< >) are non-terminal symbols, while variables not enclosed are terminal symbols. Non-terminal symbols on the left hand side of a "::=" mark can be expanded to symbols on the right hand side. The "|" mark means selection; a symbol can be expanded into any symbols connected by the "|" mark.

```
<building> ::= <base> <main> <roofs>
<base> ::= wall floor | wall | floor
<main> ::= beam <mainlist> beam
<mainlist> ::= window | door |
<mainlist> <mainlist> |
<mainlist> beam <mainlist>
<roofs> ::= <rooflist> | toproof |
<rooflist> toproof
<rooflist> ::= roof | roof <rooflist>
```

### 3.2. Various Styles

Architectural styles in this study imitate Chinese and Japanese building styles in real world. Chinese style and Japanese style share similar beam, floor and wall, but their toproof, roof, window and door are different from each other. In particular, in our work, there is only 1 type of beam shared by both Chinese style and Japanese style.

### 3.3. An Example of Rules and Generated Levels

---

[1] https://github.com/lucasnfe/AngryBirdsCover (last accessed on Feb 7, 2016)

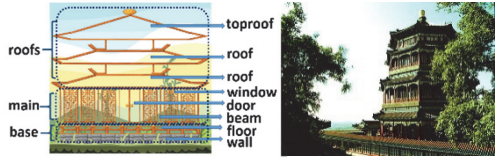

Figure 1: A Chinese style building generated (left) and a similar building in real world (right)[2]

Below is an example of a rule written in 2D-BCG. A level generated by the example rule and a real building similar to the generated building are shown in Figure 1.

<building> ::= <base> <main> <roofs>
<base> ::= wall floor
<main> ::= beam window window beam window door
           window beam window window beam
<roofs> ::= roof roof toproof

To maintain consistency within the structure of the generated building, the same style is used for all parts of the same element in the structure. For example, if the building has two windows, the same style will be selected and applied to both windows.

### 3.4. Generated Levels

While 2D-BCG can generate infinite sets of rules, in this study only five rules (Table 1) were considered. All rules were created based on the structures of several historical buildings in China and Japan. The number of models available for each element is shown in Table 2.

Table 1: Rules and models for both styles

| C | wall | floor | window | door | roof | toproof | J | wall | floor | window | door | roof | toproof |
|---|---|---|---|---|---|---|---|---|---|---|---|---|---|
| Rule1 |  |  | √ | √ | √ | √ | Rule1 | √ |  | √ | √ | √ |  |
| Rule2 | √ |  | √ | √ |  | √ | Rule2 | √ |  | √ | √ | √ |  |
| Rule3 | √ | √ | √ |  | √ |  | Rule3 | √ |  | √ | √ |  |  |
| Rule4 | √ |  | √ | √ | √ |  | Rule4 | √ |  | √ | √ |  |  |
| Rule5 | √ |  | √ | √ | √ |  | Rule5 | √ |  | √ | √ |  | √ |

Table 2: The number of models

| models | window | door | roof | toproof | common models count | wall | floor |
|---|---|---|---|---|---|---|---|
| Chinese-Style | 3 | 3 | 3 | 2 |  | 3 | 2 |
| Japanese-Style | 3 | 3 | 3 | 2 |  | 3 | 2 |

The number of all Chinese style, Japanese style and composite style levels that can be generated by our rules is 567, 540, and 10125. There were 10 rules for the composite style, which combines rules of both Chinese style and Japanese style. It shows that even with a small number of rules and models, a vast diversity of levels can be created.

### 4. Experiment

A survey was conducted to assess the visual design of the generated levels. The experiment concerned 25 voluntary participants, among which 23 were male and 2 were female, and 23 were within the age range 18 – 24 and the rest were within 25 – 34. Each subject was presented generated levels and was asked to identify whether the building is a Chinese or Japanese style. The nationalities of our participants included Chinese(1), French(2), Japanese(19), Thai(1), Vietnamese(1), and Other(1).

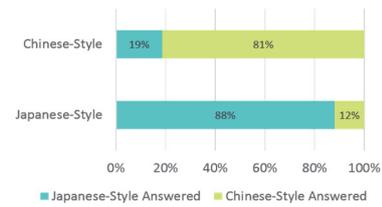

Figure 2: Summary of survey results

In order to test whether the style of generated level is recognizable to the users, a survey was conducted on Survey Monkey. The survey included an information page, a hint page, a quiz page and a survey page. Each participant had to input his/her gender, age and nationality. Features of Chinese architectural style and Japanese style were explained on the hint page. Photos of actual buildings, windows, doors, roofs and armaments were displayed along with explanatory text. The participant could access the hint page at any time.

The quiz page tested whether the participant had understood the difference between Chinese architectural style and Japanese style. He or she was shown 10 pictures and asked if the building in the picture is Chinese style or Japanese style. The participant must answer all questions correctly before moving on to the survey page. On the survey page, the participant was presented 3 Chinese style generated buildings and 3 Japanese style generated buildings randomly, and for each level, the subject had to identify whether it was Chinese style or Japanese style.

Responses to the survey are illustrated in Figure 2, from which the participants correctly identified both Chinese style (about 81%) and Japanese style building (88%). In terms of $\chi^2$-distribution, the percentages of correct answers for both styles are significant, individually compared to 50%, at 5% significance level; and there is no significant difference between both styles. As such, the architectural styles of generated levels are recognizable to users.

### 5. Conclusions

This paper presented a new approach for generating a visually attractive building that imitates Chinese style, Japanese style or a composite style architecture in the Angry Birds game. A conducted experiment confirmed that levels that mimic Chinese style or Japanese style architecture can be generated procedurally. In our future study, other architectural styles will be considered.